\def\year{2019}\relax
\documentclass[letterpaper]{article} %DO NOT CHANGE THIS
\usepackage{aaai19}  %Required
\usepackage{times}  %Required
\usepackage{helvet}  %Required
\usepackage{courier}  %Required
\usepackage{url}  %Required
\usepackage{graphicx}  %Required
\usepackage{amssymb}
\usepackage{amsmath}
\usepackage{algorithm, algorithmic}

\usepackage{booktabs}  
\usepackage{threeparttable}
\usepackage{multirow}

\begin{document}
\title{Dependency or Span, End-to-End Uniform Semantic Role Labeling}
%\author{AAAI Press\\
%Association for the Advancement of Artificial Intelligence\\
%2275 East Bayshore Road, Suite 160\\
%Palo Alto, California 94303\\
%}
\author{Zuchao Li$^{1,2,}$\thanks{$\ $ These authors made equal contribution.$^{\dag}$ Corresponding author. This paper was partially supported by National Key Research and Development Program of China (No. 2017YFB0304100), National Natural Science Foundation of China (No. 61672343 and No. 61733011), Key Project of National Society Science Foundation of China (No. 15-ZDA041), The Art and Science Interdisciplinary Funds of Shanghai Jiao Tong University (No. 14JCRZ04).}$\ $, Shexia He$^{1,2,*}$, Hai Zhao$^{1,2,\dag}$, Yiqing Zhang$^{1,2}$, Zhuosheng Zhang$^{1,2}$, 
	\\  \Large\textbf{Xi Zhou$^{3}$, Xiang Zhou$^{3}$}\\
	$^{1}$Department of Computer Science and Engineering, Shanghai Jiao Tong University \\
	$^{2}$Key Laboratory of Shanghai Education Commission for Intelligent Interaction \\ and Cognitive Engineering, Shanghai Jiao Tong University, Shanghai, China\\
	$^{3}$CloudWalk Technology, Shanghai, China\\
	{\tt \{charlee,heshexia\}@sjtu.edu.cn, zhaohai@cs.sjtu.edu.cn,}\\
	{\tt  \{zhangyiqing,zhangzs\}@sjtu.edu.cn, \{zhouxi,zhouxiang\}@cloudwalk.cn}\\
}
\maketitle
\begin{abstract}
Semantic role labeling (SRL) aims to discover the predicate-argument structure of a sentence. End-to-end SRL without syntactic input has received great attention. However, most of them focus on either span-based or dependency-based semantic representation form and only show specific model optimization respectively. Meanwhile, handling these two SRL tasks uniformly was less successful. This paper presents an end-to-end model for both dependency and span SRL with a unified argument representation to deal with two different types of argument annotations in a uniform fashion. Furthermore, we jointly predict all predicates and arguments, especially including long-term ignored predicate identification subtask. Our single model achieves new state-of-the-art results on both span (CoNLL 2005, 2012) and dependency (CoNLL 2008, 2009) SRL benchmarks. 
 
\end{abstract}

\section{Introduction}
\begin{figure}
	\centering
	\includegraphics[scale=1]{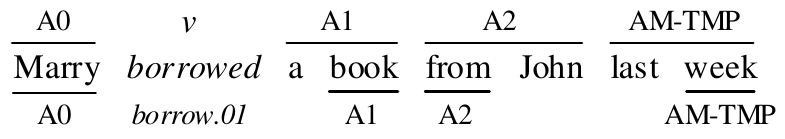}
	\caption{\label{fig:example} Examples of annotations in span (above) and dependency (below) SRL.}
\end{figure}

The purpose of semantic role labeling (SRL) is to derive the meaning representation for a sentence, which is beneficial to a wide range of natural language processing (NLP) tasks \cite{wang2016bilingual,zhang2018know}. SRL can be formed as four subtasks, including predicate detection, predicate disambiguation, argument identification and argument classification. For argument annotation, there are two formulizations. One is based on text spans, namely span-based SRL. The other is dependency-based SRL, which annotates the syntactic head of argument rather than entire argument span. Figure \ref{fig:example} shows example annotations.

%, which suffer the risk of erroneous syntactic input, leading to undesired error propagation. MoreovAer, the sophisticated tree structure of syntactic relation has made it challenging to effectively incorporate syntactic information into neural SRL models.
Great progress has been made in syntactic parsing \cite{dozat2017deep,li2018seq2seq,li2018joint}. Most traditional SRL methods rely heavily on syntactic features. To alleviate the inconvenience, recent works \cite{zhou-xu2015,marcheggiani2017,he-acl2017,selfatt2018,he2018jointly,he:2018Syntax,cai2018full} propose end-to-end models for SRL, putting syntax aside and still achieving favorable results. However, these systems focus on either span or dependency SRL,
%and lack of a uniform fashion to tackle both of them
which motivates us to explore a uniform approach. 

\begin{table*}[!htp]
	\renewcommand\arraystretch{1.3}
	\setlength{\tabcolsep}{5pt}
	\centering
	\small 
	%\footnotesize
	%\scriptsize
	%\begin{tabular}{l|l|c|c|l|c||l|l|c|c|l|c}
	\begin{tabular}{llcclc|llcclc}
		\hline
		
		\hline
		\multicolumn{6}{c|}{\textbf{Span (CoNLL 2005)}} & \multicolumn{6}{c}{\textbf{Dependency (CoNLL 2009)}} \\ 
		\hline
		\textbf{Time} & System  & SA & ST & Method & \textbf{F$_1$} & \textbf{Time} & System & SA & ST & Method & \textit{\textbf{F$_1$}} \\ 
		\hline
		\textbf{2008} & Punyakanok et al. & + & & ILP & \textbf{76.3} & \textbf{2009} & \citeauthor{Zhao2009Conll} & + &  & ME & \textit{\textbf{86.2}} \\ 
		\hline
		\textbf{2008} & Toutanova et al. & + &  & DP & \textbf{79.7} & \textbf{2010} & \citeauthor{bjorkelund2010} & + &  & global & \textit{\textbf{86.9}} \\ 
		\hline
		
		\hline
		\multicolumn{1}{l}{\textbf{2015}} & \multicolumn{1}{l}{\textbf{\citeauthor{Fitzgerald2015}}} &
		\multicolumn{1}{c}{+} & \multicolumn{1}{c}{} & \multicolumn{1}{l}{structured} & \multicolumn{1}{c}{\textbf{79.4}} & \multicolumn{2}{l}{} & \multicolumn{1}{c}{+} & \multicolumn{1}{c}{} & \multicolumn{1}{l}{structured} & \multicolumn{1}{c}{\textit{\textbf{87.3}}}\\
		\hline
		
		\hline
		\textbf{2015} & \citeauthor{zhou-xu2015} &  & +  & deep BiLSTM & \textbf{82.8} &  &  &  &  &  & \\
		\hline
		&  &  &  &  &  & \textbf{2016} & \citeauthor{roth2016} & +  &  & PathLSTM & \textit{\textbf{87.7}} \\ 
		\hline
		\textbf{2017} & \citeauthor{he-acl2017} &  & + & highway BiLSTM & \textbf{83.1} & \textbf{2017} & Marcheggiani et al. &  & +  & BiLSTM  & \textit{\textbf{87.7}} \\ 
		\hline
		&  &  &  &  &  & \textbf{2017} & \citeauthor{marcheggianiEMNLP2017} & +  & +  & GCNs  & \textit{\textbf{88.0}} \\
		\hline
		\textbf{2018} & \citeauthor{selfatt2018} &  & +  & self-attention & \textbf{84.8} & \textbf{2018} & \citeauthor{he:2018Syntax} (b) & + & +  & ELMo  & \textit{\textbf{89.5}} \\ 
		\hline
		\textbf{2018} & \citeauthor{Strubell2018}  & +  &  & self-attention & \textbf{83.9} & \textbf{2018} & \citeauthor{cai2018full} &  &  & biaffine  & \textit{\textbf{89.6}} \\ 
		\hline
		\textbf{2018} & \citeauthor{he2018jointly} (a) &  &  & ELMo  & \textbf{87.4} & \textbf{2018} & \citeauthor{li2018unified} (a) & + & + & ELMo & \textit{\textbf{89.8}} \\ 
		\hline
		
		\hline
		\textbf{2019} & \multicolumn{1}{l}{Li et al. (b) AAAI} &
		\multicolumn{1}{c}{} & \multicolumn{1}{c}{} & \multicolumn{1}{l}{ELMo+biaffine} & \multicolumn{1}{c}{\textbf{87.7}} & \multicolumn{2}{l}{} & \multicolumn{1}{c}{} & \multicolumn{1}{c}{} & \multicolumn{1}{l}{ELMo+biaffine} & \multicolumn{1}{c}{\textit{\textbf{90.4}}}\\
		\hline
		
		\hline
	\end{tabular}
	\caption{A chronicle of related work for span and dependency SRL. SA represents syntax-aware system (no + indicates syntax-agnostic system) and ST indicates sequence tagging model. F$_1$ is the result of single model on official test set.}
	%Acronyms used: SA - syntax-aware
	\label{tab:related work}
\end{table*}

Both span and dependency are effective formal representations for semantics, though for a long time it has been kept unknown which form, span or dependency, would be better for the convenience and effectiveness of semantic machine learning and later applications. Furthermore, researchers are interested in two forms of SRL models that may benefit from each other rather than their separated development. This topic has been roughly discussed in \cite{johansson2008EMNLP}, who concluded that the (best) dependency SRL system at then clearly outperformed the span-based (best) system through gold syntactic structure transformation. However, \citeauthor{johansson2008EMNLP} \shortcite{johansson2008EMNLP} like all other traditional SRL models themselves had to adopt rich syntactic features, and their comparison was done between two systems in quite different building styles.
%, which makes their conclusion still doubtful.
Instead, this work will develop full syntax-agnostic SRL systems with the same fashion for both span and dependency representation, so that we can revisit this issue under a more solid empirical basis.

In addition, most efforts focus on argument identification and classification since span and dependency SRL corpora have already marked predicate positions. Although no predicate identification is needed, it is not available in many downstream applications. Therefore, predicate identification should be carefully handled in a complete practical SRL system. To address this problem, \citeauthor{he2018jointly} \shortcite{he2018jointly} proposed an end-to-end approach for jointly predicting predicates and arguments for span SRL. Likewise, \citeauthor{cai2018full} \shortcite{cai2018full} introduced an end-to-end model to naturally cover all predicate/argument identification and classification subtasks for dependency SRL.

%Recent methods \cite{zhou-xu2015,marcheggiani2017,marcheggianiEMNLP2017,he-acl2017,selfatt2018} deal with all words in entire sentence instead of distinguishing arguments and non-arguments which actually differ in quantity. The indiscriminate treatment would result in a serious unbalanced issue for argument labeling.

%\begin{figure*}[htp]
%	\centering
%	\includegraphics[scale=0.33]{overview.pdf}
%	\caption{\label{fig:related}A general summary of related work for span-based and dependency-based SRL.}
%\end{figure*}

To jointly predict predicates and arguments, we present an end-to-end framework for both span and dependency SRL. Our model extends the span SRL model of \citeauthor{he2018jointly} \shortcite{he2018jointly}, directly regarding all words in a sentence as possible predicates, considering all spans or words as potential arguments and learning distributions over possible predicates. However, we differ by (1) introducing unified argument representation to handle two different types of SRL tasks, and (2) employing biaffine scorer to make decisions for predicate-argument relationship.

The proposed models are evaluated on span SRL datasets: CoNLL 2005 and 2012 data, as well as the dependency SRL dataset of CoNLL 2008 and 2009 shared tasks. For span SRL, our single model outperforms the previous best results by 0.3\% and 0.5\% F$_1$-score on CoNLL 2005 and 2012 test sets respectively. For dependency SRL, we achieve new state-of-the-art of 85.3\% F$_1$ and 90.4\% F$_1$ on CoNLL 2008 and 2009 benchmarks respectively. 
%Experimental results show that our model is able to improve the performance of both span and dependency SRL. To sum up, we make the following contributions.
%
%$\bullet$ We present an end-to-end syntax-agnostic SRL model\footnote{Our code is available here: \url{https://github.com/bcmi220/unisrl}.} for jointly predicting predicates and arguments.
%
%$\bullet$ We propose adaptive argument representation, which enables our model to effectively handle span-based and dependency-based SRL in a uniform fashion.
%
%$\bullet$ Our model achieves state-of-the-art results on the CoNLL 2005, 2012 and CoNLL 2008, 2009 benchmarks.

\section{Background}

%In this section, we briefly describe the SRL task and provide a review of previous and related work. SRL is a shallow semantic parsing task
SRL is pioneered by \citeauthor{gildea2002} \shortcite{gildea2002}, which uses the PropBank conventions \cite{propbank}.
%, be it span or dependency based
%In this section, we briefly describe the SRL task and provide a review of previous and related work. SRL is a shallow semantic parsing task pioneered by \cite{gildea2002}, be it span or dependency based, which can be defined as automatic derivation of meaning representation, e.g. \textit{who} did \textit{what} to \textit{whom}, \textit{where} and \textit{when}, etc. It is worth noting that SRL uses the PropBank conventions \cite{propbank}.
%\paragraph{Span SRL}
Conventionally, span SRL consists of two subtasks, argument identification and classification. The former identifies the arguments of a predicate, and the latter assigns them semantic role labels, namely, determining the relation between arguments and predicates. The PropBank defines a set of semantic roles to label arguments, falling into two categories: \textit{core} and \textit{non-core} roles. The core roles (A0-A5 and AA) indicate different semantics in predicate-argument structure, while the non-core roles are modifiers (AM-\textit{adj}) where \textit{adj} specifies the adjunct type, such as temporal (AM-TMP) and locative (AM-LOC) adjuncts. For example shown in Figure \ref{fig:example}, A0 is a proto-agent, representing the \textit{borrower}.
%Given a sentence \textit{Marry borrowed a book from John last week} with a known predicate \textit{borrowed}, Figure \ref{fig:example} shows results of argument labeling (e.g., A0 is a proto-agent, representing the \textit{borrower}). 

%In concept, the former recognizes whether a word is a true argument of a predicate, and the latter determines the relation between the argument and predicate, also known as semantic role. Given a sentence with a known predicate, Figure \ref{fig:example} shows example results of argument labeling. 

%\paragraph{Dependency SRL}

%It is worth noting that span-based SRL generally assumes gold predicates as input. 

Slightly different from span SRL in argument annotation, dependency SRL labels the syntactic heads of arguments rather than phrasal arguments, which was popularized by CoNLL-2008 and CoNLL-2009 shared tasks\footnote{CoNLL-2008 is an English-only task, while CoNLL-2009 extends to a multilingual one. Their main difference is that predicates have been beforehand indicated for the latter. Or rather, CoNLL-2009 does not need predicate identification, but it is an indispensable subtask for CoNLL-2008.} \cite{surdeanu-EtAl2008,hajivc-EtAl2009}. Furthermore, when no predicate is given, two other indispensable subtasks of dependency SRL are predicate identification and disambiguation. One is to identify all predicates in a sentence, and the other is to determine the senses of predicates. As the example shown in Figure \ref{fig:example}, \textit{01} indicates the first sense from the PropBank sense repository for predicate \textit{borrowed} in the sentence.

\subsection{Related Work}

%In early work of SRL, most of the researchers focus on feature engineering based on training corpus.
The traditional approaches on SRL were mostly about designing hand-crafted feature templates and then employ linear classifiers such as \cite{pradhan2005,punyakanok2008importance,Zhao2009Conll}. Even though neural models were introduced, early work still paid more attention on syntactic features.
%Especially with the impressive success of neural networks, considerable attention has been paid to syntactic features.
For example, \citeauthor{Fitzgerald2015} \shortcite{Fitzgerald2015} integrated syntactic information into neural networks with embedded lexicalized features, while \citeauthor{roth2016} \shortcite{roth2016} embedded syntactic dependency paths between predicates and arguments. Similarly, \citeauthor{marcheggianiEMNLP2017} \shortcite{marcheggianiEMNLP2017} leveraged the graph convolutional network to encode syntax for dependency SRL. Recently, \citeauthor{Strubell2018} \shortcite{Strubell2018} presented a multi-task neural model to incorporate auxiliary syntactic information for SRL, \citeauthor{li2018unified} \shortcite{li2018unified} adopted several kinds of syntactic encoder for syntax encoding while \citeauthor{he:2018Syntax} \shortcite{he:2018Syntax} used syntactic tree for argument pruning.

However, using syntax may be quite inconvenient sometimes, %syntactic parsers are unreliable on out-of-domain data, 
recent studies thus have attempted to build SRL systems without or with little syntactic guideline. \citeauthor{zhou-xu2015} \shortcite{zhou-xu2015} proposed the first syntax-agnostic model for span SRL using LSTM sequence labeling,
%an end-to-end model for span SRL without relying on syntactic features, 
while \citeauthor{he-acl2017} \shortcite{he-acl2017} further enhanced their model using highway bidirectional LSTMs with constrained decoding. Later, \citeauthor{selfatt2018} \shortcite{selfatt2018} presented a deep attentional neural network for applying self-attention to span SRL task. Likewise for dependency SRL, \citeauthor{marcheggiani2017} \shortcite{marcheggiani2017} proposed a syntax-agnostic model with effective word representation and obtained favorable results. \citeauthor{cai2018full} \shortcite{cai2018full} built a full end-to-end model with biaffine attention and outperformed the previous state-of-the-art.

More recently, joint predicting both predicates and arguments has attracted extensive interest on account of the importance of predicate identification, including %some recent models
\cite{he-acl2017,Strubell2018,he2018jointly,cai2018full} and this work.
%, we extend existing approaches to jointly predict all predicates and arguments
In our preliminary experiments, we tried to integrate the self-attention into our model,
%combine our model with the self-attention
but it does not provide any significant performance gain on span or dependency SRL, which is not consistent with the conclusion in \cite{selfatt2018} and lets us exclude it from this work. 

\begin{figure*}
	\centering
	\includegraphics[scale=0.82]{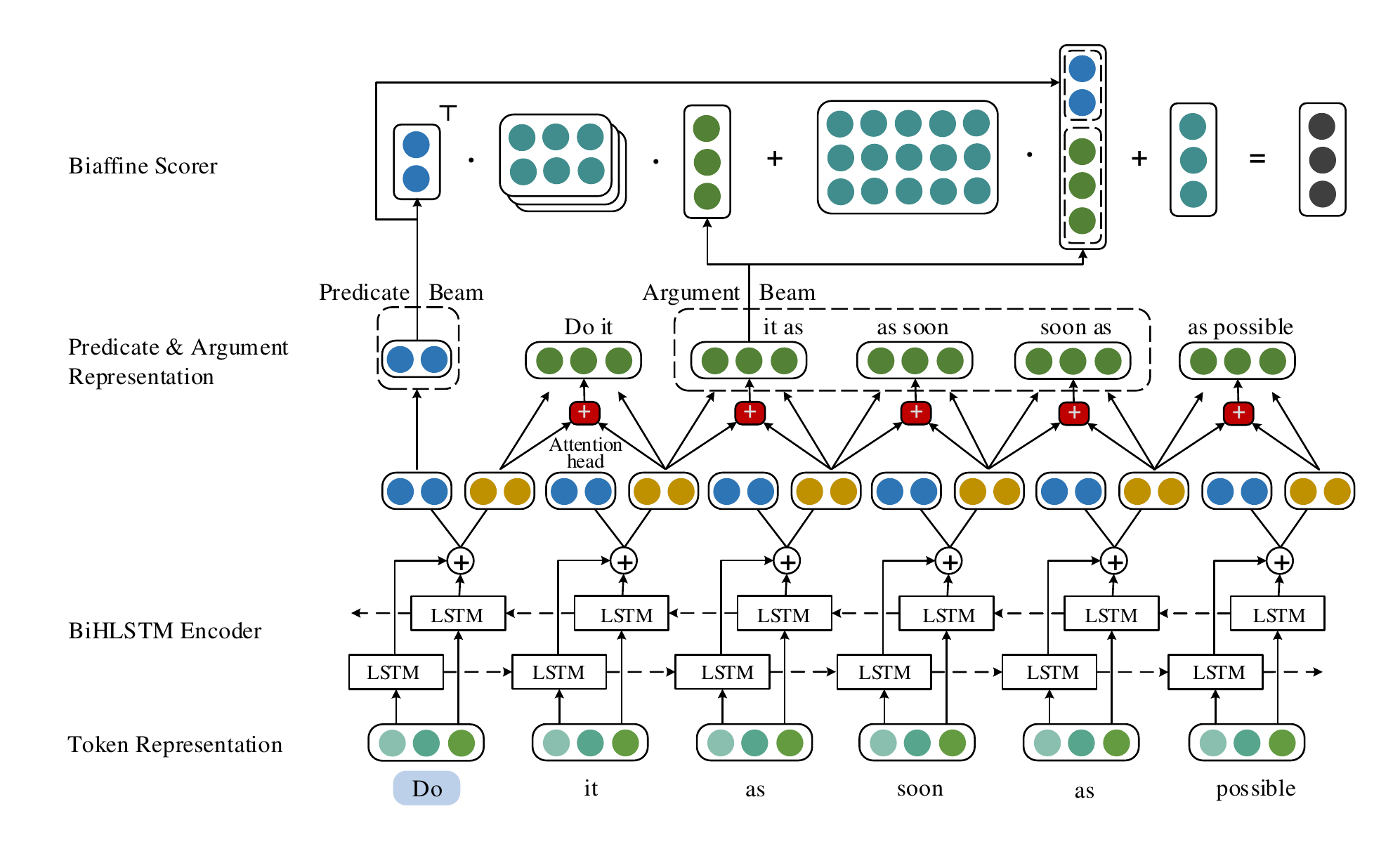}
	\caption{\label{fig:overview} The framework of our end-to-end model for uniform SRL.}
\end{figure*}

Generally, the above work is summarized in Table \ref{tab:related work}. Considering motivation, our work is most closely related to the work of \citeauthor{Fitzgerald2015} \shortcite{Fitzgerald2015}, which also tackles span and dependency SRL in a uniform fashion. The essential difference is that their model employs the syntactic features and takes pre-identified predicates as inputs, while our model puts syntax aside and jointly learns and predicts predicates and arguments. %Different from most existing SRL systems, we introduce span-ranking approach and encode each sentence only once with span-graph formulation.

%Unlike previous methods, apart from \cite{Strubell2018,he2018jointly}, 

\section{Uniform End-to-End Model}

\subsection{Overview}

%We propose an extension to the span SRL model of \citeauthor{he2018jointly} \shortcite{he2018jointly} to perform uniform SRL task. 
Given a sentence $s=w_1, w_2, \dots, w_n$, we attempt to predict a set of predicate-argument-relation tuples $\mathcal{Y} \in \mathcal{P} \times \mathcal{A} \times \mathcal{R}$, where $\mathcal{P} = \{w_1, w_2, ..., w_n\}$ is the set of all possible predicate tokens, $\mathcal{A} = \{(w_i,\dots,w_j) | 1 \leq i \leq j \leq n \}$ includes all the candidate argument spans or dependencies\footnote{When $i$=$j$, it means span degrades to dependency.}, and $\mathcal{R}$ is the set of the semantic roles. To simplify the task, we introduce a null label $\epsilon$ to indicate no relation between arbitrary predicate-argument pair following \citeauthor{he2018jointly} \shortcite{he2018jointly}. As shown in Figure \ref{fig:overview}, our uniform SRL model includes four main modules: 

$\bullet$ token representation component to build token representation $x_i$ from word $w_i$,

$\bullet$ a BiHLSTM encoder that directly takes sequential inputs,

$\bullet$ predicate and argument representation module to learn candidate representations,

$\bullet$ a biaffine scorer which takes the candidate representations as input and predicts semantic roles.

%(1) token representation component to build token representation $x_i$ from word $w_i$, (2) a BiLSTM encoder that directly takes sequential inputs, (3) predicate and argument representation module to learn candidate representations, and (4) a biaffine scorer which takes the candidate representations as input and predicts semantic roles.

\subsection{Token Representation}
We follow the bi-directional LSTM-CNN architecture \cite{chiu2016named}, where convolutional neural networks (CNNs) encode characters inside a word $w$ into character-level representation $w_{char}$ then concatenated with its word-level $w_{word}$ into context-independent representation. To further enhance the word representation, we leverage an external representation $w_{elmo}$ from pretrained ELMo (Embeddings from Language Models) layers according to \citeauthor{ELMo} \shortcite{ELMo}. 
%ELMo is obtained by deep bidirectional language model that takes characters as input, enriching subword information and contextual information, which has expressive representation power. 
Eventually, the resulting token representation is concatenated as 
%$repr_{token} = [repr_{char}, repr_{word}, repr_{elmo}]$.
\begin{equation}
x = [w_{char} , w_{word} , w_{elmo}].\nonumber
\end{equation}
%As previous dependency SRL work \cite{marcheggiani2017,he:2018Syntax} has demonstrated that the POS tag, lemma and predicate-specific feature\footnote{We use the predicate-specific feature only when the predicate is beforehand indicated.} are helpful in promoting the role labeling process, we use these features only for dependency SRL to make the results comparable.

\subsection{Deep Encoder}

The encoder in our model adopts the bidirectional LSTM with highway connections (BiHLSTM) %\cite{zhang2016highway} 
to contextualize the representation into task-specific representation: 
%which we denote as $: 
$x^c_i \in X^c: X^c = \textit{BiHLSTM}(\{x_i\}) $,
%\begin{equation}
%X^c = \textit{Bi-HLSTM}(repr_{token})
%\end{equation}
where the gated highway connections is used to alleviate the vanishing gradient problem when training very deep BiLSTMs.% which is implemented with transform gates to control the weight of linear and non-linear transformations between layers.

\subsection{Predicate and Argument Representation}

We employ contextualized representations for all candidate arguments and predicates. As referred in \cite{dozat2017deep}, applying a multi-layer perceptron (MLP) to the recurrent output states before the classifier has the advantage of stripping away irrelevant information for the current decision. Therefore, to distinguish the currently considered predicate from its candidate arguments in SRL context, we add an MLP layer to contextualized representations for argument $g^a$ and predicate $g^p$ candidates specific representations respectively with ReLU \cite{nair2010rectified} as its activation function:
%\begin{equation}
%g^a = \textit{ReLU}(\textit{MLP}_a (X^c))
%%g^p = \textit{ReLU}(\textit{MLP}_p (X^c)) 
%\end{equation}
%\begin{equation}
%%g^a = \textit{ReLU}(\textit{MLP}_a (X^c))
%g^p = \textit{ReLU}(\textit{MLP}_p (X^c))
%\end{equation}
$$g^a = \textit{ReLU}(\textbf{MLP}_a (X^c))$$
$$g^p = \textit{ReLU}(\textbf{MLP}_p (X^c))$$

To  perform uniform SRL, we introduce unified argument representation. For dependency SRL, we assume single word argument span by limiting the length of candidate argument to be 1, so our model uses the $g^a$ as the final argument representation $g^a_f$ directly. While for span SRL, we utilize the approach of span representation from \citeauthor{lee2017end} \shortcite{lee2017end}. Each candidate span representation $g^a_f$ is built by
%we build it from several aspects, 
%the boundary representation, the specific notion of headedness and the size of span: $g^a_f = g^a_{\textit{START}} \odot g^a_{\textit{END}} \odot h_\lambda \odot \phi(\lambda)$,
\begin{equation}
g^a_f = [g^a_{\textit{START}} , g^a_{\textit{END}} , h_\lambda , size(\lambda)],\nonumber
\end{equation}
where $g_{START}^a$  and $g_{END}^a$ are boundary representations, $\lambda$ indicates a span, $size(\lambda)$ is a feature vector encoding the size of span, and $h_\lambda$ is the specific notion of headedness which is learned by attention mechanism \cite{bahdanau2014neural} over words in each span (where $t$ is the position inside span) as follows :
%\begin{equation}
$$\mu^a_t = \textbf{w}_{attn} \cdot \textbf{MLP}_{attn} (g^a_t)$$ 
%\end{equation}
%\begin{equation}
$$\nu_t = \frac{\exp(\mu^a_t)}{\sum_{k=\textit{START}}^{\textit{END}} \exp(\mu^a_k)}$$
%a_t = \frac{\exp(g^a_t)}{\sum_{k=\textit{START}}^{\textit{END}} \exp(g^a_k)}
%h_s = \sum_{t=\textit{START}}^{\textit{END}} a_t \cdot g^a_t
%\end{equation}
%\begin{equation}
$$h_{\lambda} = \sum_{t=\textit{START}}^{\textit{END}} \nu_t \cdot g^a_t$$
%a_t = \frac{\exp(g^a_t)}{\sum_{k=\textit{START}}^{\textit{END}} \exp(g^a_k)}
%h_\lambda = \sum_{t=\textit{START}}^{\textit{END}} a_t \cdot g^a_t
%\end{equation}

\subsection{Scorers}%Biaffine 
For predicate and arguments, we introduce two unary scores on their candidates:
$$\phi_p = \textbf{w}_p \textbf{MLP}^s_p(g^p),$$
$$\phi_a = \textbf{w}_a \textbf{MLP}^s_a(g_f^a).$$
For semantic role,
%between predicates and arguments
we adopt a relation scorer with biaffine attention \cite{dozat2017deep}:
%\begin{equation}
%e_t = \{g^p_t\}^T \textbf{W} g^a_f + \textbf{U}^T g^p_t + \textbf{V}^T g^a_f + \textbf{b}
%\end{equation}
\begin{align}
\Phi_{r}(p, a) &= \textit{Biaffine}(g^p, g^a_f)\nonumber\\
	 &= \{g^p_t\}^T \textbf{W}_1 g^a_f \label{bilinear}\\
	&+ \textbf{W}_2^T (g^p_t \oplus g^a_f) + \textbf{b} \label{hescorer}
\end{align}
%and feed-forward networks scorer \cite{he2018jointly}:
%\begin{equation}
%\Phi_{r}(p, a) = \textbf{w}_{r}^{T}\textbf{MLP}_{r} ([g^p; g^a_f])
%%e_t = \{g^p_t\}^T \textbf{W} g^a_f + \textbf{U}^T g^p_t + \textbf{V}^T g^a_f + \textbf{b}
%\end{equation}
%$e_i$ indicates the scores of each relation between candidate predicate and argument,A
where $\textbf{W}_1$ and $\textbf{W}_2$ respectively denote the weight matrix of the bi-linear and the linear terms and $\textbf{b}$ is the bias item.

The biaffine scorer differs from feed-forward networks scorer in bilinear transformation. Since SRL can be regarded as a classification task, the distribution of classes is uneven and the problem comes worse after the null labels are introduced. The output layer of the model normally includes a bias term designed to capture the prior probability of each class, with the rest of the model focusing on learning the likelihood of every classes occurring in data. 
%To address such uneven problem, \citeauthor{dozat2017deep} \shortcite{dozat2017deep} introduced the bias terms into the bilinear attention, resulting in a biaffine transformation. 
The biaffine attention as Dozat and Manning (2017) in our model directly assigns a score for each specific semantic role and would be helpful for semantic role prediction. Actually, (He et al., 2018a) used a scorer as Equation (2), which is only a part of our scorer including both Equations (\ref{bilinear}) and (\ref{hescorer}). Therefore, our scorer would be more informative than previous models such as \cite{he2018jointly}.
%Therefore, our scorer contains more information than feed-forward networks based on the predicate and argument representations.

\begin{table*}
	\centering
	\begin{tabular}{lccccccccc}  
		\toprule  
		\multirow{2}{*}{End-to-End}&\multicolumn{3}{c}{CoNLL-2005 WSJ}&\multicolumn{3}{c}{CoNLL-2005 Brown}&\multicolumn{3}{c}{CoNLL-2012 (OntoNotes)}\cr  
		\cmidrule(lr){2-4} \cmidrule(lr){5-7} \cmidrule(lr){8-10}
		&P&R&F$_1$&P&R&F$_1$&P&R&F$_1$\cr  
		\midrule  
		\citeauthor{he-acl2017} \shortcite{he-acl2017} \small{(Single)} &80.2&82.3&81.2&67.6&69.6&68.5&78.6&75.1&76.8\cr
		\citeauthor{Strubell2018} \shortcite{Strubell2018} &83.7&83.7&83.7&72.6&69.7&71.1&80.7&79.1&79.9\cr
		\citeauthor{he2018jointly} \shortcite{he2018jointly} &84.8&87.2&86.0&73.9&\textbf{78.4}&76.1&81.9&\textbf{84.0}&82.9\cr  
		\textbf{Ours} \small{(Single)}& \textbf{85.2}&\textbf{87.5}&\textbf{86.3}&\textbf{74.7}&78.1&\textbf{76.4}&\textbf{84.9}&81.4&\textbf{83.1} \cr
		\midrule  
		\citeauthor{he-acl2017} \shortcite{he-acl2017} \small{(Ensemble)} &82.0&83.4&82.7&69.7&70.5&70.1&80.2&76.6&78.4\cr  
		\bottomrule  
	\end{tabular}
	\caption{End-to-end span SRL results on CoNLL-2005 and CoNLL-2012 data, compared with previous systems in terms of precision (P), recall (R), F$_1$-score. The CoNLL-2005 contains two test sets: WSJ (in-domain) and Brown (out-of-domain).}\label{tab:end-for-span}
\end{table*}

\subsection{Training Objective}

The model is trained to optimize the probability $\textit{P}_\theta(\hat{y}|s)$ of the predicate-argument-relation tuples $\hat{y}_{(p,a,r)} \in \mathcal{Y}$ given the sentence $s$, which can be factorized as:
\begin{align}
\textit{P}_\theta(y|s) &= \prod_{p \in \mathcal{P},a \in \mathcal{A},r \in \mathcal{R}} \textit{P}_\theta(y_{(p,a,r)}|s)\nonumber\\
&= \prod_{p \in \mathcal{P},a \in \mathcal{A},r \in \mathcal{R}} \frac{\phi(p, a,r)}{\sum_{\hat{r} \in \mathcal{R}}\phi(p, a,\hat{r})}\nonumber
\end{align}
where $\theta$ represents the model parameters, and $\phi(p, a, r) = \phi_p + \phi_a + \Phi_{r}(p, a)$, is the score for the predicate-argument-relation tuple, including predicate score $\phi_p$, argument score $\phi_a$ and relation score $\Phi_{r}(p, a)$.
%\begin{equation}
%\alpha(p, a, r) = \alpha_p + \alpha_a + \Phi_{r}(p, a)
%\end{equation}

%\paragraph{Predicate \& Argument Prediction}
%\paragraph{Prediction}
%\noindent \textbf{Prediction} \quad
Our model adopts a biaffine scorer for semantic role label prediction, which is implemented as cross-entropy loss. Moreover, our model is trained to minimize the negative likehood of the golden structure $y$: $\mathcal{J}(s) = -\log \textit{P}_\theta(y|s)$.
%\begin{equation}
%\mathcal{J}(s) = -\log \textit{P}_\theta(y|s)
%\end{equation}
The score of null labels are enforced into $\phi(p, a, \epsilon) = 0$. For predicates and arguments prediction, we train separated scorers ($\phi_p$ and $\phi_a$) in parallel fed to the biaffine scorer for predicate and argument predication respectively, which helps to reduce the chance of error propagation. 

%\paragraph{Label Prediction}

\subsection{Candidates Pruning}

The number of candidate arguments for a sentence of length $n$ is $\textit{O}(n^2)$ for span SRL, and $\textit{O}(n)$ for dependency. As the model deals with $\textit{O}(n)$ possible predicates, the computational complexity is $\textit{O}(n^3 \cdot |\mathcal{R}|)$ for span, $\textit{O}(n^2 \cdot |\mathcal{R}|)$  for dependency, which is too computationally expensive.%impractical.

To address this issue, we attempt to prune candidates using two beams for storing the candidate arguments and predicates with size $\beta_pn$ and $\beta_an$ inspired by \citeauthor{he2018jointly} \shortcite{he2018jointly}, where $\beta_p$ and $\beta_a$ are two manually setting thresholds. First, the predicate and argument candidates are ranked according to their predicted score ($\phi_p$ and $\phi_a$) respectively, and then we reduce the predicate and argument candidates with defined beams. Finally, we take the candidates from the beams to participate the label prediction. Such pruning will reduce the overall number of candidate tuples to  $\textit{O}(n^2 \cdot |\mathcal{R}|)$ for both types of tasks. Furthermore, for span SRL, we set the maximum length of candidate arguments to $\mathcal{L}$, which may decrease the number of candidate arguments to $\textit{O}(n)$.

\begin{table}
	\centering
	%\small
	\setlength{\tabcolsep}{4pt}
	\begin{tabular}{lcccccc}
		\toprule  
		\multirow{2}{*}{Systems}&\multicolumn{3}{c}{WSJ}&\multicolumn{3}{c}{Brown}\\  
		\cmidrule(lr){2-4} \cmidrule(lr){5-7}
		&P&R&F$_1$&P&R&F$_1$ \\  
		\midrule
		J \& N \shortcite{Johansson2008Dependency} &$-$&$-$&81.75&$-$&$-$&69.06\\
%		Zhao \shortcite{Zhao2008Parsing} & $-$ & $-$ & 77.67 & $-$ & $-$ & $-$ \\
		Zhao et al. \shortcite{zhao2009} & $-$ & $-$ & 82.1 & $-$ & $-$ & $-$ \\
		Zhao et al. \shortcite{zhao-jair-2013} & $-$ & $-$ & 82.5 & $-$ & $-$ & $-$ \\
		\citeauthor{he:2018Syntax} \shortcite{he:2018Syntax} &83.9&82.7&83.3&$-$&$-$&$-$\\
		\citeauthor{cai2018full}\shortcite{cai2018full}&\textbf{84.7}&85.2&85.0&$-$&$-$&72.5\\
		\textbf{Ours} &84.5&\textbf{86.1}&\textbf{85.3}&74.6&73.8&\textbf{74.2} \\  
		\bottomrule
	\end{tabular}
	\caption{Dependency SRL results on CoNLL-2008 test sets.}\label{tab:end-for-dependency}
\end{table}

\subsection{SRL Constraints}
%Following the definition of the semantic roles, there are some dependencies between them. Since the predicted predicate-argument-relation tuples from the model may violate the SRL constraints described in
According to PropBank semantic convention, predicate-argument structure has to follow a few of global constraints \cite{punyakanok2008importance,he-acl2017}, we thus incorporate constraints on the output structure with a dynamic programing decoder during inference. These constraints are described as follows:

$\bullet$ Unique core roles (U): Each core role (A0-A5, AA) should appear at most once for each predicate.

$\bullet$ Continuation roles (C): A continuation role C-X can exist only when its base role X is realized before it.

$\bullet$ Reference roles (R): A reference role R-X can exist only when its base role X is realized (not necessarily before R-X).

$\bullet$ Non-overlapping (O): The semantic arguments for the same predicate do not overlap in span SRL.

%Note that the C and R constraints are more commonly violated in gold data, which results in worse performance. Therefore
As C and R constraints lead to worse performance in our models from our preliminary experiments, we only enforce U and O constraints on span SRL and U constraints on dependency SRL\footnote{O constraint will be automatically satisfied for dependency, as it may be regarded as length 1 sized span.}.

\begin{table*}
	\centering
	\begin{tabular}{llccccccccc}  
		\toprule  
		\multicolumn{2}{c}{\multirow{2}{*}{System}}&\multicolumn{3}{c}{CoNLL-2005 WSJ}&\multicolumn{3}{c}{CoNLL-2005 Brown}&\multicolumn{3}{c}{CoNLL-2012 (OntoNotes)}\cr  
		\cmidrule(lr){3-5} \cmidrule(lr){6-8} \cmidrule(lr){9-11}& &P&R&F$_1$&P&R&F$_1$&P&R&F$_1$\cr  
		\midrule
		\multirow{10}{*}{Single} & \citeauthor{CoNLL2013} \shortcite{CoNLL2013} \small{(Revised)} &$-$&$-$&$-$&$-$&$-$&$-$&78.5&76.6&77.5\cr
%		&T{\"a}ckstr{\"o}m et al. \shortcite{tackstrom2015} \small{(Struct.)}&82.3&77.6&79.9&74.3&68.6&71.3&80.6&78.2&79.4\cr 
		& \citeauthor{zhou-xu2015} \shortcite{zhou-xu2015}&82.9&82.8&82.8&70.7&68.2&69.4&$-$&$-$&81.3\cr
%		& \citeauthor{yang2017A} \shortcite{yang2017A}&$-$&$-$&81.9&$-$&$-$&72.0&$-$&$-$&$-$ \cr  
		& \citeauthor{he-acl2017} \shortcite{he-acl2017}&83.1&83.0&83.1&72.9&71.4&72.1&81.7&81.6&81.7\cr
		& \citeauthor{selfatt2018} \shortcite{selfatt2018}&84.5&85.2&84.8&73.5&74.6&74.1&81.9&83.6&82.7\cr
		& \citeauthor{ELMo} \shortcite{ELMo}&$-$&$-$&$-$&$-$&$-$&$-$&$-$&$-$&84.6\cr
		& \citeauthor{Strubell2018} \shortcite{Strubell2018}&83.9&83.9&83.9&73.3&71.8&72.5&$-$&$-$&$-$\cr  
		& \citeauthor{he2018jointly} \shortcite{he2018jointly}&$-$&$-$&87.4&$-$&$-$&80.4&$-$&$-$&85.5\cr  
		& \textbf{Ours} & \textbf{87.9} & \textbf{87.5} & \textbf{87.7} & \textbf{80.6} & \textbf{80.4} & \textbf{80.5} &\textbf{85.7} &\textbf{86.3} &\textbf{86.0} \cr
		\midrule
		\multirow{5}{*}{Ensemble} & Punyakanok et al. \shortcite{punyakanok2008importance}&82.3&76.8&79.4&73.4&62.9&67.8&$-$&$-$&$-$ \cr
		& Toutanova et al. \shortcite{toutanova2008global}&81.9&78.8&80.3&$-$&$-$&68.8&$-$&$-$&$-$ \cr
		& \citeauthor{Fitzgerald2015} \shortcite{Fitzgerald2015}&82.5&78.2&80.3&74.5&70.0&72.2&81.2&79.0&80.1\cr  
		&\citeauthor{he-acl2017} \shortcite{he-acl2017}&85.0&84.3&84.6&74.9&72.4&73.6&83.5&83.3&83.4\cr
		&\citeauthor{selfatt2018} \shortcite{selfatt2018}&85.9&86.3&86.1&74.6&75.0&74.8&83.3&84.5&83.9 \cr  
		\bottomrule  
	\end{tabular}
	\caption{Span SRL results with pre-identified predicates on CoNLL-2005 and CoNLL-2012 test sets.}\label{tab:gold-for-span}
\end{table*}

\section{Experiments}
Our models\footnote{Our code is available here: \url{https://github.com/bcmi220/unisrl}.} are evaluated on two PropBank-style SRL tasks: span and dependency. For span SRL, we test model on the common span SRL datasets from CoNLL-2005 \cite{CoNLL2005} and CoNLL-2012 \cite{CoNLL2013} shared tasks. For dependency SRL, we experiment on CoNLL 2008 \cite{surdeanu-EtAl2008} and 2009 \cite{hajivc-EtAl2009} benchmarks. As for the predicate disambiguation in dependency SRL task, we follow the previous work \cite{roth2016}.

We consider two SRL setups: \textit{end-to-end} and \textit{pre-identified predicates}. 
%In the \textit{end-to-end}
For the former setup, our system jointly predicts all the predicates and their arguments in one shot, which turns into CoNLL-2008 setting for dependency SRL. In order to compare with previous models, we also report results with \textit{pre-identified predicates}, where predicates have been beforehand identified in corpora. Therefore, the experimental results fall into two categories: end-to-end results and results with pre-identified predicates. 
%We present all results using the official evaluation script from the CoNLL-2005 and CoNLL-2009 shared task.

\subsection{Datasets}
%Our model is evaluated on the CoNLL-2005, CoNLL-2009, CoNLL-2012 shared task, following the standard training, development and test splits.

%\paragraph{CoNLL 2005 and 2012}
\noindent \textbf{CoNLL 2005 and 2012} \quad The CoNLL-2005 shared task focused on verbal predicates only for English.
%and is only concerned with the English language. 
The CoNLL-2005 dataset takes section 2-21 of Wall Street Journal (WSJ) data as training set, and section 24 as development set. The test set consists of section 23 of WSJ for in-domain evaluation together with 3 sections from Brown corpus for out-of-domain evaluation. The larger CoNLL-2012 dataset is extracted from OntoNotes v5.0 corpus, which contains both verbal and nominal predicates.

%CoNLL-2005 \cite{CoNLL2005} and CoNLL-2012 (OntoNotes 5.0 \cite{CoNLL2013}) are two commonly-used PropBank-style, span-based SRL dataset with gold predicates provided. We use the official evaluation script from the CoNLL-2005 shared task for evaluation on both datasets. For CoNLL-2005 shared task, we experiment on both the in-domain(WSJ) and out-of-domain(Brown) test sets to make results comparable.
%\paragraph{CoNLL 2008 and 2009}

\noindent \textbf{CoNLL 2008 and 2009} \quad
CoNLL-2008 and the English part of CoNLL-2009 shared tasks use the same English corpus, 
%The CoNLL-2009 shared task is dedicated to dependency-based SRL in multiple languages, 
which merges two treebanks, PropBank and NomBank. NomBank is a complement to PropBank with similar semantic convention for nominal predicate-argument structure annotation. Besides, the training, development and test splits of English data are identical to that of CoNLL-2005.

\subsection{Setup}
\paragraph{Hyperparameters}
%In our experiments, the word embeddings are 300-dimensional GloVe vectors pre-trained on 840B tokens \cite{penningtonEMNLP2014} and updated during training. The character representations are randomly initialized, and the dimension is 8. In the character CNN, the convolutions  have window sizes of 3, 4, and 5 characters, each consisting of 50 filters. Besides, we use 3 stacked bidirectional LSTMs with 200 dimensional hidden states. Each MLP consists of two hidden layers with 150 dimensions and takes ReLU as activation function. For candidates pruning, we follow the settings of \cite{he2018jointly}, modeling spans up to length 30 for span SRL and 1 for dependency SRL, using $\lambda_p=0.4$ for pruning predicates and $\lambda_a=0.8$ for pruning arguments.

In our experiments, the word embeddings are 300-dimensional GloVe vectors \cite{penningtonEMNLP2014}. The character representations with dimension 8 randomly initialized. In the character CNN, the convolutions  have window sizes of 3, 4, and 5, each consisting of 50 filters. Moreover, we use 3 stacked bidirectional LSTMs with 200 dimensional hidden states. The outputs of BiLSTM employs two 300-dimensional MLP layers with the ReLU as activation function.
%for learning specific representation of predicates and arguments
Besides, we use two 150-dimensional hidden MLP layers with ReLU to score predicates and arguments respectively. For candidates pruning, we follow the settings of \citeauthor{he2018jointly} \shortcite{he2018jointly}, modeling spans up to length $\mathcal{L}=30$ for span SRL and $\mathcal{L}=1$ for dependency SRL, using $\beta_p=0.4$ for pruning predicates and $\beta_a=0.8$ for pruning arguments.

%\paragraph{Training Details}

\noindent \textbf{Training Details} \quad
During training, we use the categorical cross-entropy as objective, with Adam optimizer \cite{adam2015} initial learning rate 0.001. We apply 0.5 dropout to the word embeddings and character CNN outputs and 0.2 dropout to all hidden layers and feature embeddings. In the LSTMs, we employ variational dropout masks that are shared across timesteps \cite{gal2016}, with 0.4 dropout rate. All models are trained for up to 600 epochs with batch size 40 on a single NVIDIA GeForce GTX 1080Ti GPU, which occupies 8 GB graphic memory and takes 12 to 36 hours.

%All models are trained for at most 320,000 steps with early stopping on the development set, which takes less than 48 hours on a single Titan X GPU. we train models for up to 600 epochs with batch size 40 on 1 NVIDIA GeForce GTX 1080Ti GPU with 1.67 steps per second and occupying 8 GB graphic memory each model, which takes 36 to 48 hours.

\subsection{End-to-end Results}
We present all results using the official evaluation script from the CoNLL-2005 and CoNLL-2009 shared tasks, and compare our model with previous state-of-the-art models.
 
%\paragraph{Span SRL}
\noindent \textbf{Span SRL}\quad Table \ref{tab:end-for-span} shows results on CoNLL-2005 in-domain (WSJ) and out-of-domain (Brown) test sets, as well as the CoNLL-2012 test set (OntoNotes). The upper part of table presents results from single models. Our model outperforms the previous models with absolute improvements in F$_1$-score of 0.3\% on CoNLL-2005 benchmark. Besides, our single model performs even much better than all previous ensemble systems.
%On all datasets, our model is able to predict over 40\% of the sentences completely correctly

%\paragraph{Dependency SRL}
\noindent \textbf{Dependency SRL}\quad Table \ref{tab:end-for-dependency} presents the results on CoNLL-2008. J \& N (2008b) \cite{Johansson2008Dependency} was the highest ranked system in CoNLL-2008 shared task. We obtain comparable results with the recent state-of-the-art method \cite{cai2018full}, and our model surpasses the model \cite{he:2018Syntax} by 2\% in F$_1$-score. 
%These results demonstrate that our model can be adapted to perform dependency SRL and achieve impressive performance gains.

\subsection{Results with Pre-identified Predicates}

\begin{table*}[t]
	\centering
	\setlength{\tabcolsep}{10pt}
	\begin{tabular}{llcccccc}
		\toprule  
		\multicolumn{2}{c}{\multirow{2}{*}{System}}&\multicolumn{3}{c}{CoNLL-2009 WSJ}&\multicolumn{3}{c}{CoNLL-2009 Brown}\\  
		\cmidrule(lr){3-5} \cmidrule(lr){6-8}
		&&P&R&F$_1$&P&R&F$_1$ \\  
		\midrule
		\multirow{10}{*}{Single} & \citeauthor{Zhao2009Conll} \shortcite{Zhao2009Conll} &$-$&$-$&86.2&$-$&$-$&74.6 \\
%		&\citeauthor{bjorkelund2010} \shortcite{bjorkelund2010} \small{(Global)} &88.6&85.2&86.9&77.9&73.6&75.7 \\
%		&\citeauthor{Lei2015} \shortcite{Lei2015} &$-$&$-$&86.6&$-$&$-$&75.6 \\
		&\citeauthor{Fitzgerald2015} \shortcite{Fitzgerald2015} \small{(Struct.)} &$-$&$-$&87.3&$-$&$-$&75.2 \\
		&\citeauthor{roth2016} \shortcite{roth2016} \small{(Global)} &90.0&85.5&87.7&78.6&73.8&76.1 \\
		&Marcheggiani et al. \shortcite{marcheggiani2017}  &88.7&86.8&87.7&79.4&76.2&77.7 \\
		&\citeauthor{marcheggianiEMNLP2017} \shortcite{marcheggianiEMNLP2017}&89.1&86.8&88.0&78.5&75.9&77.2 \\
		&\citeauthor{he:2018Syntax} \shortcite{he:2018Syntax} &89.7&89.3&89.5&\textbf{81.9}&76.9&79.3 \\
		&\citeauthor{cai2018full} \shortcite{cai2018full} &89.9&89.2&89.6&79.8&78.3&79.0 \\
		&\citeauthor{li2018unified} \shortcite{li2018unified} &  90.3 & 89.3 & 89.8 &  80.6 & 79.0 & 79.8 \\
		&\textbf{Ours}&89.6&\textbf{91.2}&\textbf{90.4}&81.7&\textbf{81.4}&\textbf{81.5}\\
		\midrule
		\multirow{3}{*}{Ensemble}
		&\citeauthor{Fitzgerald2015} \shortcite{Fitzgerald2015} &$-$&$-$&87.7&$-$&$-$&75.5 \\
		&\citeauthor{roth2016} \shortcite{roth2016} &90.3&85.7&87.9&79.7&73.6&76.5 \\
		&\citeauthor{marcheggianiEMNLP2017} \shortcite{marcheggianiEMNLP2017} &\textbf{90.5}&87.7&89.1&80.8&77.1&78.9 \\ 
		\bottomrule
	\end{tabular}
	\caption{Dependency SRL results with pre-identified predicates on CoNLL-2009 English benchmark.}\label{tab:gold-for-dependency}
\end{table*}

To compare with to previous systems with pre-identified predicates, we report results from our models as well.

%\paragraph{Span SRL}
\noindent \textbf{Span SRL}\quad 
%We report the comparison of performance between our model and previous work in 
Table \ref{tab:gold-for-span} shows that our model outperforms all published systems, even the ensemble model \cite{selfatt2018}, achieving the best results of 87.7\%, 80.5\% and 86.0\% in F$_1$-score respectively. %Furthermore, our approach provides significantly improved performance by concatenating the contextualized ELMo representations to tokens.

%\paragraph{Dependency SRL}

\noindent \textbf{Dependency SRL}\quad Table \ref{tab:gold-for-dependency} compares the results of dependency SRL on CoNLL-2009 English data. Our single model gives a new state-of-the-art result of 90.4\% F$_1$ on WSJ. For Brown data, the proposed syntax-agnostic model yields a performance gain of 1.7\% F$_1$ over the syntax-aware model \cite{li2018unified}.%, which makes use of syntactic features. 

\subsection{Ablation}
To investigate the contributions of ELMo representations and biaffine scorer in our end-to-end model, we conduct a series of ablation studies on the CoNLL-2005 and CoNLL-2008 WSJ test sets, unless otherwise stated.

\begin{table}
	\centering
	\begin{tabular}{lccc}
		\toprule
		System & CoNLL-2005& CoNLL-2008\\
		\midrule
		\citeauthor{cai2018full} \shortcite{cai2018full} & $-$ & 85.0  \\
		\midrule
		\citeauthor{he2018jointly} \shortcite{he2018jointly} & 86.0 & $-$ \\
		w/o ELMo & 82.5 & $-$ \\ 
		\midrule
		Ours & 86.3 & 85.3 \\
		w/o ELMo & 83.0 & 85.1 \\
		w/o biaffine scorer & 85.8 & 83.7 \\
		\bottomrule
	\end{tabular}
	\caption{Effectiveness of ELMo representations and biaffine scorer on the CoNLL 2005 and 2008 WSJ sets.}\label{ablation:elmo}
\end{table}

%\paragraph{Effectiveness of ELMo representations}
Table \ref{ablation:elmo} compares F$_1$ scores of \citeauthor{he2018jointly} \shortcite{he2018jointly} and our model without ELMo representations. We observe that effect of ELMo is somewhat surprising, where removal of the ELMo dramatically declines the performance by 3.3-3.5 F$_1$ on CoNLL-2005 WSJ. However, our model gives quite stable performance for dependency SRL regardless of whether ELMo is concatenated or not. The results indicate that ELMo is more beneficial to span SRL.

%\begin{table}
%	\centering
%	\setlength{\tabcolsep}{4pt}
%	\begin{tabular}{lccc}
%		\toprule
%		System & CoNLL-2005& CoNLL-2008\\
%		\midrule
%		\citeauthor{he2018jointly} \shortcite{he2018jointly} & 86.0 & $-$ \\
%		w/o ELMo & 82.5 & $-$ \\ 
%		\midrule
%		Ours & 86.3 & 85.3 \\
%		w/o ELMo & ? & 85.1 \\
%		\bottomrule
%	\end{tabular}
%	\caption{Contribution of ELMo representations on the CoNLL 2005 and 2008 WSJ sets.}\label{ablation:elmo}
%\end{table}
%
%\begin{table}
%	\centering
%	\setlength{\tabcolsep}{4pt}
%	\begin{tabular}{lccc}
%		\toprule
%		System & CoNLL-2005& CoNLL-2008\\
%		\midrule
%		\citeauthor{he2018jointly} \shortcite{he2018jointly} & 86.0 & $-$ \\
%		\citeauthor{cai2018full} \shortcite{cai2018full} & $-$ & 85.0  \\ 
%		\midrule
%		Ours & 86.3 & 85.3 \\
%		w/o biaffine scorer & 85.8 & ? \\
%		\bottomrule
%	\end{tabular}
%	\caption{Contribution of biaffine scorer on the CoNLL 2005 and 2008 WSJ sets.}\label{ablation:biaffine}
%\end{table}

%\paragraph{Effectiveness of biaffine scorer}
In order to better understand how the biaffine scorer influences our model performance, we train our model with different scoring functions. To ensure a fair comparison with the model \cite{he2018jointly}, we replace the biaffine scorer with their scoring functions implemented with feed-forward networks, and the results of removing biaffine scorer are also presented in Table \ref{ablation:elmo}. We can see 0.5\% and 1.6\% F$_1$ performance degradation on CoNLL 2005 and 2008 WSJ respectively. The comparison shows that the biaffine scorer is more effective for scoring the relations between predicates and arguments. Furthermore, these results show that biaffine attention mechanism is applicable to span SRL. 

\subsection{Dependency or Span?}
%\noindent \textbf{Dependency-based Evaluation} \quad
%we convert the span-style arguments to the dependency-style arguments using gold dependency syntax. If the syntactic structure of the span is a subtree, we take the root node of the subtree as the dependency SRL argument; if not,  (C): we take the root node of the deepest subtree in the forest ; (C-E): we remove the span-style argument when the syntactic structure is forest. Table \ref{tab:convert} shows corresponding results.
It is very hard to say which style of semantic formal representation, dependency or span, would be more convenient for machine learning as they adopt incomparable evaluation metric. Recent researches \cite{peng2018learning} have proposed to learn semantic parsers from multiple datasets in Framenet style semantics, while our goal is to compare the quality of different models in the span and dependency SRL for Propbank style semantics. Following \citeauthor{johansson2008EMNLP} \shortcite{johansson2008EMNLP}, we choose to directly compare their performance in terms of dependency-style metric through a transformation way. 
%Thus we need to convert span SRL results into dependency form. 
Using the head-finding algorithm in \cite{johansson2008EMNLP} which used gold-standard syntax, we may determine a set of head nodes for each span. This process will output an upper bound performance measure about the span conversion due to the use of gold syntax.

We do not train new models for the conversion and the resulted comparison. Instead, we do the job on span-style CoNLL 2005 test set and dependency-style CoNLL 2009 test set (WSJ and Brown), considering these two test sets share the same text content. As the former only contains verbal predicate-argument structures, for the latter, we discard all nomial predicate-argument related results and predicate disambiguation results during performance statistics. Table \ref{tab:convert} shows the comparison.

On a more strict setting basis, the results from our same model for  span and dependency SRL verify the same conclusion of \citeauthor{johansson2008EMNLP} \shortcite{johansson2008EMNLP}, namely, dependency form is in a favor of machine learning effectiveness for SRL even compared to the conversion upper bound of span form.
%Note our results come from more strict setting than \citeauthor{johansson2008EMNLP} \shortcite{johansson2008EMNLP}.

%\begin{table}
%	\centering
%	\setlength{\tabcolsep}{4pt}
%	\begin{tabular}{llll}
%		\toprule
%		WSJ & Dep F$_1$ & Span-converted F$_1$ & $\Delta$ F \\
%		\midrule
%		J \& N & 85.93 & 84.32 & 1.61 \\
%		Our system & 90.16 & 89.2 (89.0) & 0.96 (1.16) \\
%		\hline
%		\\
%		\midrule
%		WSJ+Brown & Dep F$_1$ & Span-converted F$_1$ & $\Delta$ F \\
%		\midrule
%		J \& N & 84.29 & 83.45 & 0.84 \\
%		Our system & 88.81 & 88.23 (88.08) & 0.58 (0.73) \\
%		\bottomrule
%	\end{tabular}
%	\caption{Dependency vs. Span-converted Dependency on CoNLL-2005 test set with dependency evaluation.}\label{tab:convert}
%\end{table}

\begin{table}
	\centering
	\setlength{\tabcolsep}{4pt}
	\begin{tabular}{clccc}
		\toprule
		& &  Dep F$_1$ & Span-converted F$_1$ & $\Delta$ F$_1$ \\
		\midrule
		 \multirow{2}{25pt}{WSJ} & J \& N & 85.93 & 84.32 & 1.61 \\
		 & Our system & 90.41 & 89.20 & 1.21 \\
		\hline
		\midrule
		\multirow{2}{25pt}{WSJ+\\Brown} & J \& N & 84.29 & 83.45 & 0.84 \\
		& Our system & 88.91 & 88.23 & 0.68 \\
		\bottomrule
	\end{tabular}
	\caption{Dependency vs. Span-converted Dependency on CoNLL 2005, 2009 test sets with dependency evaluation.}\label{tab:convert}
\end{table}

%\begin{table}
%	\centering
%	\setlength{\tabcolsep}{4pt}
%	\begin{tabular}{lccc}
%		\toprule
%		System & P & R & F$_1$ \\
%		\midrule
%		 \citeauthor{johansson2008EMNLP} \shortcite{johansson2008EMNLP}& 88.46 & 88.0 & 85.93 \\
%		\midrule
%		C & 88.8 & 89.4 & 89.0 \\
%		C-E & 90.08 & 88.48 & 89.2 \\
%		Our 09 Model & 90.04 &  90.27 & 90.16  \\
%		\bottomrule
%	\end{tabular}
%	\caption{Comparison of CoNLL-2005 shared task with dependency-based evaluation.}\label{ablation:gap}
%\end{table}

\section{Conclusion}
This paper presents an end-to-end neural model for both span and dependency SRL, which may jointly learn and predict all predicates and arguments. We extend existing model and introduce unified argument representation with biaffine scorer to the uniform SRL for both span and dependency representation forms. Our model achieves new state-of-the-art results on the CoNLL 2005, 2012 and CoNLL 2008, 2009 benchmarks. Our results show that span and dependency SRL can be effectively handled in a uniform fashion, which for the first time enables us to conveniently explore the useful connection between two types of semantic representation forms. 

\bibliographystyle{aaai} 
\bibliography{references}

\end{document}